\documentclass[letterpaper, 10 pt, conference]{ieeeconf}

\IEEEoverridecommandlockouts
\usepackage{graphicx}
\usepackage{amsmath}
\usepackage{amssymb}
\usepackage{booktabs}
\usepackage{multirow}
\usepackage{caption}
\usepackage{cite}
\usepackage{url}
\usepackage[hidelinks,bookmarks=false]{hyperref}

\title{\LARGE \bf
UMI-Bridge: Action-Anchored Latent Alignment across \\ Human and Robot Manipulation Data
}

\author{
\authorblockN{Haiyi Liu$^{1}$ \quad Jinming Ma$^{2,\dagger,\ddagger}$ \quad Ke Rui$^{2}$ \quad Yuteng Wei$^{2}$ \quad Yuan Ma$^{2}$ \\
Yushen Zuo$^{2}$ \quad Honglong Tian$^{2}$ \quad Haoran Jia$^{2}$ \quad Weitao Zhou$^{2}$ \quad Jiawei Wang$^{2}$ \\
Shiyi Chen$^{1}$ \quad Haiyan Mao$^{1}$ \quad Jiaqi Zhang$^{1}$ \quad Chun Zhang$^{1}$ \quad Minglei Li$^{2,\ddagger}$}
\authorblockA{$^{1}$Tsinghua University \qquad $^{2}$Simple AI\\
\textnormal{Project page: \href{https://umi-bridge.github.io/}{https://umi-bridge.github.io/}}\\
$^{\dagger}$Project lead. \quad $^{\ddagger}$Corresponding authors: \textnormal{\{majinming, liminglei\}@simpleai.tech}}
}

\IEEEaftertitletext{%
\vspace{-1.5\baselineskip}%
\begin{minipage}{\textwidth}
  \centering
  \captionsetup{skip=4pt}
  \includegraphics[width=0.96\textwidth]{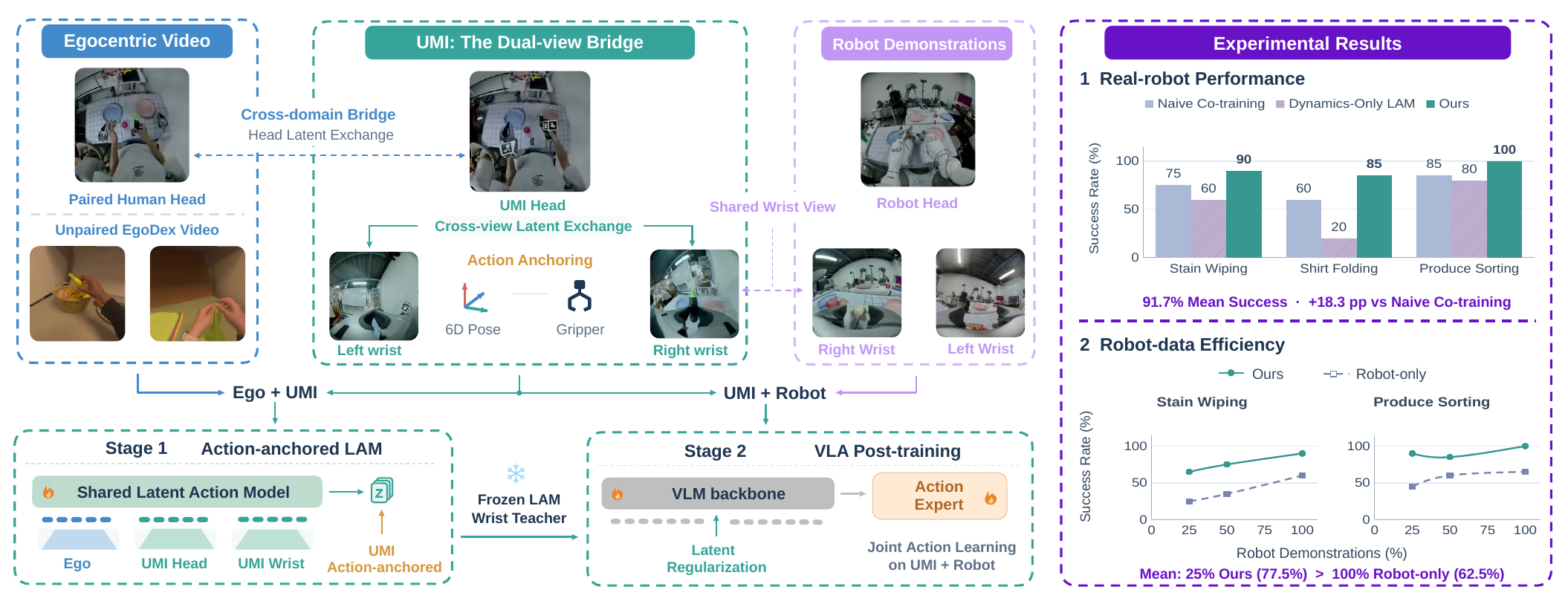}
  \captionof{figure}{\textbf{UMI-Bridge overview.}
  UMI-Bridge learns action-anchored latents from human manipulation data using UMI action supervision and alignment across views and domains.
  Stage~1 uses egocentric videos and UMI demonstrations; Stage~2 applies frozen wrist-latent and dynamics supervision during VLA post-training on UMI and robot data.
  Right: real-robot performance and data efficiency, with $20$ rollouts per bar or point.
  Means cover the tasks shown; the full-data Ours results are reused across panels.}
  \label{fig:concept}
  \vspace{0pt}
\end{minipage}%
}

\hypersetup{
  pdftitle={UMI-Bridge: Action-Anchored Latent Alignment across Human and Robot Manipulation Data},
  pdfauthor={Haiyi Liu; Jinming Ma; Ke Rui; Yuteng Wei; Yuan Ma; Yushen Zuo; Honglong Tian; Haoran Jia; Weitao Zhou; Jiawei Wang; Shiyi Chen; Haiyan Mao; Jiaqi Zhang; Chun Zhang; Minglei Li},
  pdfsubject={Action-anchored latent alignment for robot manipulation}
}

\begin{document}

\maketitle
\thispagestyle{empty}
\pagestyle{empty}

\begin{abstract}
Real-robot demonstrations are limited, motivating the use of human manipulation data collected without robots, including egocentric videos and handheld Universal Manipulation Interface (UMI) demonstrations.
However, differences in viewpoint, embodiment, and available action supervision make it difficult to align representations across these sources according to manipulation motion rather than visual appearance.
We introduce UMI-Bridge, which uses UMI as an intermediate domain to align representations according to action equivalence rather than pixel similarity.
UMI action supervision anchors the latent representation to end-effector motion and gripper behavior, while synchronized head--wrist observations and paired ego--UMI clips support alignment across views and domains.
We train a dual-view latent action model (LAM) on human manipulation data without robot demonstrations, then freeze its wrist teacher and dynamics model to regularize vision-language-action (VLA) post-training on UMI and robot data.
The shared wrist interface enables this training-time supervision across both domains while preserving the policy's standard inference architecture.
Across three real-robot tasks, UMI-Bridge achieves $91.7\%$ mean success versus $73.3\%$ for Naive Co-training with matched UMI and robot data.
On two data-efficiency tasks, it surpasses a full-data Robot-only baseline using $25\%$ of the robot demonstrations together with UMI data.
It also achieves $85\%$ and $90\%$ success on two additional tasks learned from UMI demonstrations without task-specific robot demonstrations.
These results support action-anchored latent alignment for data-efficient robot learning and UMI-to-robot task transfer.
\end{abstract}

\section{Introduction}
\label{sec:intro}

Training vision-language-action (VLA) models for robot manipulation commonly relies on demonstrations that connect visual observations and task instructions to executable actions~\cite{kim2024openvla,intelligence2025pi05}.
Acquiring these demonstrations requires robot access and operator effort, constraining data collection for new tasks~\cite{khazatsky2024droid}.
Human manipulation data, including egocentric videos and demonstrations collected with the Universal Manipulation Interface (UMI), offer additional manipulation experience without requiring robot operation~\cite{grauman2022ego4d,hoque2025egodex,chi2024umi}.
However, these sources differ from robot demonstrations in viewpoint, embodiment, and available action supervision.
Incorporating them into VLA training therefore requires representations that connect the manipulation experience they contain to the observations and actions used for robot control.

A central challenge is determining which information should be shared across these domains.
Corresponding manipulation motions can produce different visual changes when performed by a human hand or a robot gripper and observed from different viewpoints.
Latent action models make unlabeled videos useful for policy learning~\cite{ye2025lapa}, with task-centric modeling and cross-view reconstruction improving the relevance of learned representations~\cite{bu2025univla,cho2026mvplam}.
However, predicting visual transitions alone does not ensure that a latent captures the motion needed for control~\cite{cdlam2025}.
Our guiding principle is therefore to \emph{align what matters for action, not what matches in pixels}.
We pursue \emph{action equivalence}: corresponding manipulation observations should admit a consistent interpretation in terms of end-effector motion and gripper behavior despite differences in appearance.

UMI~\cite{chi2024umi} provides an intermediate domain for learning action-relevant representations from human manipulation data.
HiFi-UMI~\cite{wei2026hifiumi} provides synchronized head and wrist observations with action labels.
The head view supports correspondence with egocentric videos, while synchronized head--wrist views enable cross-view alignment.
UMI action supervision anchors the latent action representation to end-effector motion and gripper behavior.
The wrist view provides a shared interface for applying the learned representation to UMI and robot policy training (Fig.~\ref{fig:concept}).

We introduce \textbf{UMI-Bridge}, a two-stage framework for learning from human manipulation data and robot demonstrations.
Stage~1 learns a dual-view latent action model (LAM) from egocentric videos and UMI demonstrations, combining feature dynamics, UMI action supervision, and alignment across views and domains; no robot demonstrations are used.
Stage~2 post-trains $\pi_{0.5}$~\cite{intelligence2025pi05} on UMI and robot data, using the frozen wrist teacher and dynamics model to supervise the policy representation.
The auxiliary latent predictions do not condition the action expert, and the LAM and auxiliary prediction head are omitted at deployment, preserving the standard VLA inference architecture.

Our experiments connect representation learning to real-robot performance.
The Dynamics-Only LAM teacher achieves better feature prediction but worse action decoding and lower policy success than the full teacher, showing that visual predictability alone is insufficient for effective policy supervision.
Across three tasks, UMI-Bridge achieves $91.7\%$ mean success, compared with $73.3\%$ for Naive Co-training using the same UMI and robot data.
On two data-efficiency tasks, our method uses only $25\%$ of the robot demonstrations, together with UMI data, to achieve $77.5\%$ mean success, exceeding the $62.5\%$ of a Robot-only policy trained on the full robot dataset.
It achieves $85\%$ and $90\%$ success on two transfer tasks for which only UMI demonstrations are available, without task-specific robot demonstrations.

Our contributions are:
\begin{enumerate}
\item \textbf{Action-anchored LAM.} A dual-view LAM trained on human manipulation data, using UMI action supervision and paired observations to learn action-relevant representations with cross-view and cross-domain correspondence.
\item \textbf{LAM-regularized VLA post-training.} A training-time regularization scheme that uses a frozen action-anchored LAM to provide wrist-latent and dynamics supervision during UMI--robot co-training, preserving the standard inference architecture.
\item \textbf{Real-robot validation.} Representation ablations and real-robot evaluations that support the full teacher design and demonstrate improved task success, robot-demonstration efficiency, and transfer from UMI demonstrations without task-specific robot data.
\end{enumerate}

\section{Related Work}
\label{sec:related}

\subsection{Vision-Language-Action Models}
VLA models transfer pretrained vision-language representations to robot control, as exemplified by RT-2 and OpenVLA~\cite{brohan2023rt2,kim2024openvla}.
The $\pi_0$ architecture introduces flow-matching action generation~\cite{black2024pi0}, while $\pi_{0.5}$ incorporates heterogeneous co-training to improve generalization~\cite{intelligence2025pi05}.
Our work focuses on data-efficient post-training, combining UMI and robot action supervision with representation supervision learned from human manipulation data.

\subsection{Latent Actions and Representation Alignment}
Latent action models learn transition representations from videos without robot action labels~\cite{schmidt2024lapo,bruce2024genie}.
LAPA uses discrete latent actions for VLA pretraining~\cite{ye2025lapa}, while UniVLA incorporates language and feature-space modeling for task-centric transfer~\cite{bu2025univla}.
CD-LAM identifies action-irrelevant visual bias in reconstruction-trained latents~\cite{cdlam2025}, and MVP-LAM reduces viewpoint dependence through cross-viewpoint reconstruction~\cite{cho2026mvplam}.
Learned latents can also supervise policy representations.
LARA jointly optimizes a LAM and a VLA through representation alignment~\cite{lara2026}; WALA combines a frozen latent encoder with a trainable dynamics decoder during policy learning~\cite{wala2026}.
UMI-Bridge first learns a dual-view LAM from human manipulation data with UMI action supervision, then freezes both its encoder and dynamics model to regularize VLA post-training.
The auxiliary latent predictions do not condition the action expert.

\subsection{Learning from Human Manipulation Data}
Human-to-robot transfer requires reconciling observation and action differences across embodiments.
EgoMimic combines cross-domain alignment with human--robot co-training~\cite{kareer2024egomimic}, while other methods use shared wrist-translation representations~\cite{chen2026bridgingaction} or robot-format pseudo-actions extracted from human video~\cite{aceego2026}.

UMI supports portable demonstration collection and direct robot transfer through a relative-trajectory action interface~\cite{chi2024umi}.
EgoGuide incorporates synchronized head and wrist views~\cite{egoguide2026}, and HiFi-UMI demonstrates direct deployment after post-training on high-fidelity handheld data alone~\cite{wei2026hifiumi}.
BRIDGE uses state-gated experts to reconcile handheld and teleoperated supervision in contact-rich tasks~\cite{surendran2026bridge}.
UMI-Bridge uses UMI actions and paired views to ground latent representations learned from human manipulation data, then applies frozen wrist-latent and dynamics supervision during UMI--robot co-training.

\section{Problem Formulation}
\label{sec:setup-data}

We consider VLA post-training with human manipulation data---egocentric videos and handheld UMI demonstrations---and executable robot demonstrations.
Our goal is to learn action-relevant shared representations across these sources and use them to support VLA post-training under a limited robot-data budget.

\subsection{Three Data Sources and the UMI Bridge}
\label{sec:data-domains}

\textbf{Egocentric human videos.}
The dataset $\mathcal{D}_E$ contains $108.1$\,h of video ($29{,}878$ bare-handed clips) from a public EgoDex subset~\cite{hoque2025egodex}.
These videos provide diverse observations of human manipulation without action labels.
Unpaired clips contribute feature-dynamics supervision, allowing the representation to learn from motion beyond the available robot demonstrations.

\textbf{UMI demonstrations.}
The dataset $\mathcal{D}_U$ contains $135.1$\,h of demonstrations ($28{,}085$ episodes) across five bimanual tasks~\cite{chi2024umi}.
Each episode provides synchronized head and wrist observations together with action labels.
UMI provides an intermediate domain: its head view supports correspondence with egocentric videos, while its wrist views and action labels support representation supervision for UMI and robot policies.

\textbf{Robot demonstrations.}
The dataset $\mathcal{D}_R$ contains $9.8$\,h of teleoperation ($809$ episodes) across three bimanual tasks.
It provides robot observations paired with executable actions and is used only for policy post-training in Stage~2.
The representation learned in Stage~1 uses no robot demonstrations.

To establish correspondence between egocentric-video and UMI observations, we additionally construct a paired collection $\mathcal{D}_{EH}^{P}$ from these two sources.
The same operator performs each task consecutively in the same scene, first bare-handed and then with the UMI gripper, with both executions recorded from a head-mounted camera.
Dynamic time warping in visual feature space~\cite{sakoe1978dtw} aligns the recordings; one-to-one matching retains $2{,}027$ high-confidence pairs ($21{,}634$ training samples).
A shared task label alone is insufficient to establish a pair.

Table~\ref{tab:data-roles} summarizes the scale and available supervision of the three data sources.
For Stage~2, the UMI dataset contains $4{,}000$ episodes for each of the five tasks, totaling $20{,}000$ episodes ($82.6$\,h).
Two tasks have UMI demonstrations but no robot demonstrations; we report transfer evaluations on both in Sec.~\ref{sec:exp-transfer}.

\begin{table}[t]
  \centering
  \vspace*{5pt}
  \caption{\textbf{Data sources, scale, and available supervision.}}
  \label{tab:data-roles}
  \setlength{\tabcolsep}{2.5pt}
  \renewcommand{\arraystretch}{1.15}
  \footnotesize
  \begin{tabular}{@{}p{0.24\columnwidth}p{0.28\columnwidth}p{\dimexpr0.48\columnwidth-4\tabcolsep\relax}@{}}
    \toprule
    \textbf{Data source} & \textbf{Data scale} & \textbf{Available supervision} \\
    \midrule
    EgoDex ($\mathcal{D}_E$)
      & 29{,}878 clips\newline108.1\,h
      & Ego RGB video\newline No action labels \\
    \addlinespace[5pt]
    UMI ($\mathcal{D}_U$)
      & 28{,}085 episodes\newline135.1\,h
      & Head/wrist RGB video\newline Action labels \\
    \addlinespace[5pt]
    Robot ($\mathcal{D}_R$)
      & 809 episodes\newline9.8\,h
      & Head/wrist RGB video\newline Executable actions \\
    \bottomrule
  \end{tabular}
  \vspace*{-5pt}
\end{table}

\subsection{Observation and Action Representations}
\label{sec:observation-action-space}

Let $I_t^v$ denote an observation at time $t$ from view $v$.
The view index covers ego ($E$), UMI head and wrist ($UH$, $UW$), and robot head and wrist ($RH$, $RW$).
A frozen DINOv2 encoder~\cite{oquab2024dinov2} produces patch features $f_t^v=E_{\mathrm{vis}}(I_t^v)$.
For each arm, a pair of observations separated by $\Delta$ is represented by one latent token $z_t^v\in\mathbb{R}^{512}$.

UMI actions $u^U$ and robot actions $a^R$ use a common relative end-effector representation~\cite{chi2024umi}.
Each arm has a $10$D action consisting of relative translation $\Delta p\in\mathbb{R}^{3}$, continuous $6$D rotation $r_{6D}$ (the first two rows of the rotation matrix)~\cite{zhou2019continuity}, and absolute gripper opening $g\in\mathbb{R}$; the bimanual action is $20$D.
The common representation specifies consistent action semantics, while each action label supervises only observations from its own domain.

\subsection{Action Equivalence and Latent Requirements}
\label{sec:action-equivalence}

We use \emph{action equivalence} as a representation objective: observations of corresponding manipulation motion should admit a consistent action interpretation despite differences in viewpoint or visual appearance.
UMI labels anchor this interpretation to end-effector motion and gripper behavior.
For human videos without action labels, paired UMI recordings provide the correspondence through which action grounding can be learned.

The shared latent should retain distinctions that matter for execution, while view-specific scene information remains available in the observation features.
This motivates four complementary requirements for $z$.

\textbf{Action grounding.}
The latent should support decoding of end-effector motion and gripper behavior wherever action labels are available.
This requirement distinguishes control-relevant motion from other changes in the scene.

\textbf{Predictability.}
Together with the current observation features, the latent should explain the subsequent feature change.
This ensures that it describes the observed temporal transition, including for videos without action labels.

\textbf{Cross-view consistency.}
For synchronized UMI observations, a latent inferred from the head view should remain informative about wrist-view dynamics, and vice versa.
The views describe the same physical motion while retaining their own visual context.

\textbf{Cross-domain sharing.}
For paired egocentric-video and UMI clips, the latent should support a consistent interpretation of manipulation motion across the two embodiments.
UMI wrist observations then provide the interface through which this action-grounded representation can supervise robot policy learning.

Section~\ref{sec:stage1} introduces the dynamics, action-anchoring, and cross-view/cross-domain losses that implement these requirements.
Section~\ref{sec:stage2} uses the resulting wrist representation as a training-time regularizer for VLA post-training.

\section{Method}
\label{sec:method}

\begin{figure*}[t]
  \centering
  \vspace*{4pt}
  \includegraphics[width=\textwidth]{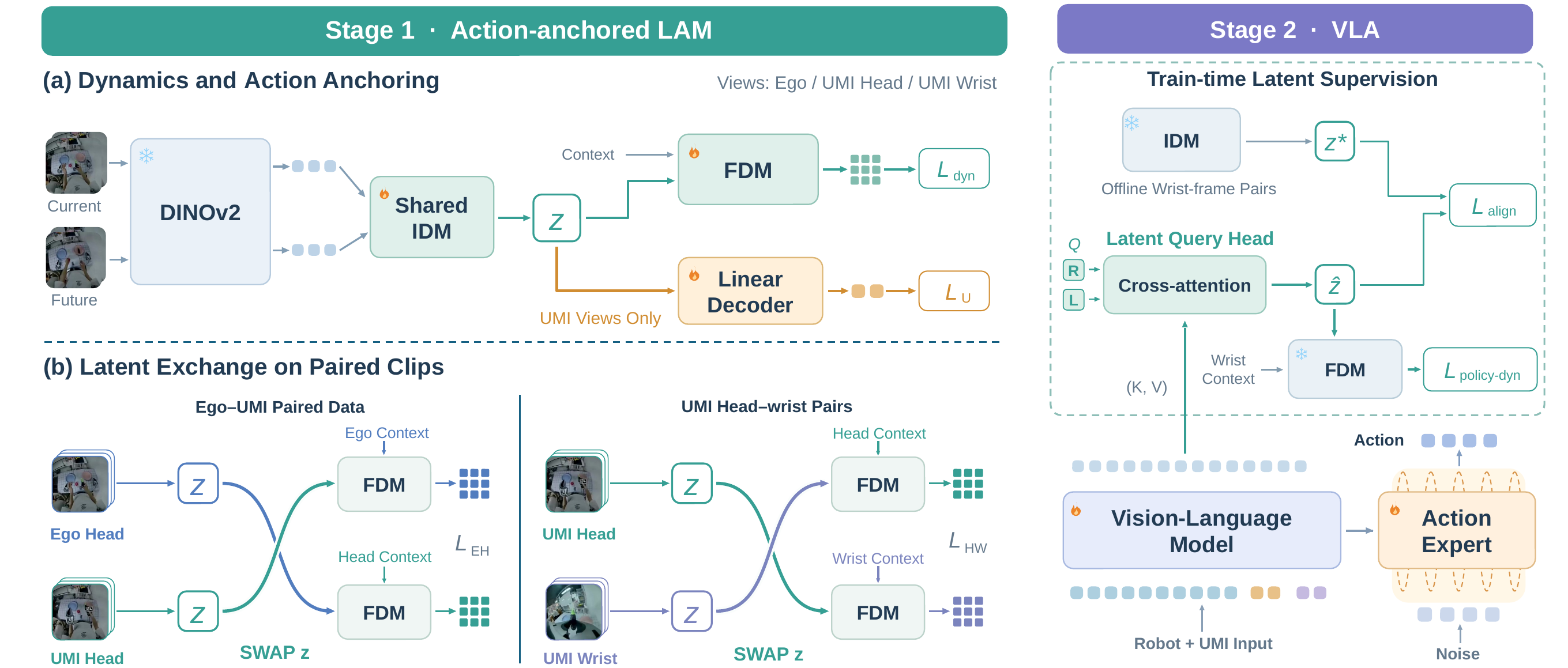}
  \par\vspace{2pt}
  \caption{\textbf{Action-anchored LAM and latent-regularized VLA post-training.}
  Stage~1 uses Ego and UMI data only: (a) a shared IDM encodes frozen DINOv2 features into latents supervised by feature dynamics $L_{\mathrm{dyn}}$ and UMI action regression $L_U$; (b) latent exchange on paired Ego--UMI head clips and synchronized UMI head--wrist clips provides $L_{EH}$ and $L_{HW}$.
  In Stage~2, arm queries attend to high-level VLM features to predict $\hat{z}$, supervised by the frozen wrist teacher through $L_{\mathrm{align}}$ and the frozen FDM through $L_{\mathrm{policy\text{-}dyn}}$.
  The dashed box encloses the additional training-only branch; $L_R$ supervises the standard action expert on both UMI and robot data.}
  \label{fig:method}
  \vspace*{-4pt}
\end{figure*}

UMI-Bridge learns action-anchored representations from human manipulation data and uses them to regularize VLA post-training on UMI and robot demonstrations (Fig.~\ref{fig:method}).
Stage~1 trains a dual-view latent action model (LAM) on egocentric videos and UMI demonstrations, combining UMI action supervision with cross-view and cross-domain alignment.
Stage~2 uses the frozen LAM's wrist latents and forward dynamics model to regularize VLA post-training on UMI and robot demonstrations.

\subsection{Stage 1: Action-Anchored Latent Alignment}
\label{sec:stage1}

\paragraph*{Shared IDM/FDM in frozen feature space}
Following the inverse/forward dynamics lineage of latent action models~\cite{schmidt2024lapo,bruce2024genie,ye2025lapa,bu2025univla}, we place both modules in frozen DINOv2 feature space rather than in pixels.
A view-agnostic inverse dynamics model (IDM) reads adjacent features,
\begin{equation}
z_t^v = Q_\phi\!\left(f_t^v,\,f_{t+\Delta}^v\right),\qquad v\in\{E,UH,UW\},
\label{eq:idm}
\end{equation}
and a forward dynamics model (FDM) predicts the future feature residual from the latent and a view embedding $e_v$,
\begin{equation}
\widehat{\Delta f}_t^v = F_\omega\!\left(f_t^v,\,z_t^v,\,e_v\right),\qquad
\Delta f_t^v = f_{t+\Delta}^v - f_t^v.
\label{eq:fdm}
\end{equation}
The IDM uses no VQ codebook or language conditioning.
Wrist images retain their fisheye projection, without a trainable visual adapter.
Freezing the LAM and visual encoder in Stage~2 fixes the teacher mapping.

\subsubsection{Feature-Space Dynamics $L_{\mathrm{dyn}}$}
With feature residual loss $\ell_f(a,b)=\|a-b\|_1+\beta\bigl(1-\cos(a,b)\bigr)$ ($\beta{=}0.2$),
\begin{equation}
L_{\mathrm{dyn}}
=\sum_{v\in\{E,UH,UW\}}
\mathbb{E}\big[\ell_f(\widehat{\Delta f}^{v},\Delta f^{v})\big],
\label{eq:ldyn}
\end{equation}
where the expectation runs over $\mathcal{D}_E$ for the ego view and over $\mathcal{D}_U$ for the two UMI views.
This objective encourages feature prediction and supplies the only supervision for unpaired ego clips.

\subsubsection{Action Anchoring $L_U$}
A dynamics-only latent may encode action-irrelevant changes, such as moving distractors~\cite{cdlam2025}.
Motion labels from the same UMI demonstrations constrain the latent toward control-relevant motion.
We regress $u^U$ from both view latents through a linear decoder $D_U$, $\hat{u}_{t:t+H}^{U,v}=D_U(z_t^v)$ for $v\in\{UH,UW\}$, and penalize each action component:
\begin{equation}
\begin{aligned}
L_U
=\sum_{v}\Big[
&\lambda_p\,\mathrm{Huber}(\widehat{\Delta p},\Delta p)
+\lambda_r\,d_{\mathrm{SO}(3)}(\hat{R},R)\\
&+\lambda_g\,\mathcal{L}_g(\hat{g},g)
\Big],
\end{aligned}
\label{eq:lu}
\end{equation}
where $\mathcal{L}_g$ is a Huber loss on gripper opening.
With a linear $D_U$, low error indicates linearly decodable action information in the latent.
Supervising synchronized head and wrist latents with the same UMI action targets anchors both representations to end-effector motion and gripper behavior.

\subsubsection{Head--Wrist Cross-View Bridge $L_{HW}$}
Anchoring alone leaves the two views free to encode motion in incompatible ways.
On synchronized UMI pairs we therefore swap latents and require the FDM to still predict the other view's residual~\cite{cho2026mvplam}:
\begin{equation}
\begin{aligned}
L_{HW}
&=
\ell_f\!\left(F_\omega(f_t^{UW},z_t^{UH},e_{UW}),\,\Delta f_t^{UW}\right)\\
&\quad+
\ell_f\!\left(F_\omega(f_t^{UH},z_t^{UW},e_{UH}),\,\Delta f_t^{UH}\right),
\end{aligned}
\label{eq:lhw}
\end{equation}
applied in both directions for head$\leftrightarrow$left-wrist and head$\leftrightarrow$right-wrist (four terms; no wrist$\leftrightarrow$wrist).
Predicting another view's transitions discourages reliance on view-specific nuisance factors.

\subsubsection{Ego--UMI-Head Cross-Domain Bridge $L_{EH}$}
We apply the same latent exchange to high-confidence cross-domain pairs $(I^E,I^{UH})\in\mathcal{D}_{EH}^{P}$:
\begin{equation}
\begin{aligned}
L_{EH}
&=
\ell_f\!\left(F_\omega(f_t^{UH},z_t^{E},e_{UH}),\,\Delta f_t^{UH}\right)\\
&\quad+
\ell_f\!\left(F_\omega(f_t^{E},z_t^{UH},e_{E}),\,\Delta f_t^{E}\right).
\end{aligned}
\label{eq:leh}
\end{equation}
Predicting target-domain feature transitions with exchanged latents encourages cross-domain compatibility.
Together with UMI action supervision, this provides an indirect route to action-relevant representations for ego observations.
Unpaired ego clips contribute only $L_{\mathrm{dyn}}$.
UMI clips provide $L_{\mathrm{dyn}}$, $L_U$, and $L_{HW}$, while paired ego--UMI samples provide $L_{EH}$.

\subsection{Stage 2: LAM-Regularized VLA Post-Training}
\label{sec:stage2}

We post-train $\pi_{0.5}$~\cite{intelligence2025pi05} on action-labeled UMI and robot demonstrations.
HiFi-UMI~\cite{wei2026hifiumi} demonstrates that high-fidelity UMI data can support VLA post-training for direct robot deployment.
Each training example pairs observations with actions from the same domain, using the common relative end-effector representation defined in Sec.~\ref{sec:observation-action-space}.

Let $c_t=(o_t,\ell_t,s_t)$ denote the policy context: visual observations, a task instruction, and proprioceptive state.
The two domains use the same state convention: end-effector pose relative to the previous frame and absolute gripper opening.
Let $A_t$ denote the corresponding action chunk after the policy's action preprocessing, constructed from $u^U$ for UMI data or $a^R$ for robot data.
Using the continuous flow-matching objective~\cite{lipman2023flow}, we interpolate $A_t^{\tau}=(1-\tau)A_t+\tau\epsilon$, where $\epsilon\sim\mathcal{N}(0,I)$ and $\tau\in[0,1]$ follows the backbone's training schedule.
The action expert predicts the velocity field:
\begin{equation}
L_R=\mathbb{E}_{\mathrm{mix},\tau,\epsilon}\!\left[
\mathrm{MSE}\!\left(v_\theta(A_t^{\tau},\tau,c_t),\epsilon-A_t\right)
\right],
\label{eq:lflow}
\end{equation}
where $\mathbb{E}_{\mathrm{mix}}$ averages over the UMI--robot training mixture and $\theta$ denotes the VLA parameters.
Here $\tau{=}1$ denotes noise and $\tau{=}0$ denotes data; $L_R$ is used for both domains.

The Stage~1 IDM $Q_\phi$, FDM $F_\omega$, and visual encoder remain frozen.
For arm $k\in\{\mathrm{left},\mathrm{right}\}$, wrist features from the same demonstration define the teacher
\begin{equation}
z_{t,k}^{\star}=\mathrm{sg}[Q_\phi(f_{t,k}^{W},f_{t+\Delta,k}^{W})]\in\mathbb{R}^{512},
\label{eq:wrist-teacher}
\end{equation}
where $\mathrm{sg}$ denotes stop-gradient.
The teacher summarizes the demonstrated motion over $\Delta$; future observations are used only to construct supervision.
This wrist interface is shared by UMI and robot examples.

To transfer this action-grounded supervision to the policy, we regularize the VLM prefix features $H_t$ computed from $c_t$.
A query head with two cross-attention layers and one learned query $q_k$ per arm predicts
$\hat{z}_{t,k}=P_\psi(H_t,q_k)\in\mathbb{R}^{512}$.
The prefix contains no future frames or action targets, and $\hat{z}_{t,k}$ is not an input to the action expert.
We apply two complementary regularizers to this prediction~\cite{yu2025repa,wala2026,lara2026}.

The alignment loss matches the action-grounded wrist teacher:
\begin{equation}
\begin{aligned}
L_{\mathrm{align}}=\mathbb{E}_{\mathrm{mix}}\!\Big[\sum_k m_{t,k}\big(
&1-\cos(\hat{z}_{t,k},z_{t,k}^{\star})\\
&+\gamma\|\hat{z}_{t,k}-z_{t,k}^{\star}\|_2^2\big)\Big],
\end{aligned}
\label{eq:lalign}
\end{equation}
where $m_{t,k}\in\{0,1\}$ indicates an active arm and $\gamma\geq0$ weights the squared-distance term.
The dynamics loss requires the predicted latent to explain the observed wrist-feature transition:
\begin{equation}
\begin{aligned}
L_{\mathrm{policy\text{-}dyn}}
=\mathbb{E}_{\mathrm{mix}}\!\Big[\sum_k m_{t,k}\,\ell_f\big(
&F_\omega(f_{t,k}^{W},\hat{z}_{t,k},e_{UW}),\\
&\Delta f_{t,k}^{W}\big)\Big],
\end{aligned}
\label{eq:lpdyn}
\end{equation}
with $\Delta f_{t,k}^{W}=f_{t+\Delta,k}^{W}-f_{t,k}^{W}$.
Both domains reuse $e_{UW}$ to access the frozen FDM through the same Stage~1 wrist interface.

Auxiliary losses update the query head and VLM prefix, while $L_R$ trains the standard VLA policy path.
Although the FDM parameters are frozen, gradients from $L_{\mathrm{policy\text{-}dyn}}$ propagate through it to $\hat{z}_{t,k}$.
The teacher targets and frozen LAM parameters receive no gradients.
The combined objective and training schedule are given in Sec.~\ref{sec:training-deployment}.

\subsection{Training and Deployment}
\label{sec:training-deployment}

Training proceeds through representation learning followed by policy adaptation.
In Stage~1, the inverse dynamics model $Q_\phi$, forward dynamics model $F_\omega$, and UMI action decoder $D_U$ are jointly optimized, while the DINOv2 encoder remains fixed.
The objective integrates feature dynamics, action grounding, and alignment across views and domains:
\begin{equation}
L_{\mathrm{S1}}
=
L_{\mathrm{dyn}}+L_U+\lambda_{HW} L_{HW}+\lambda_{EH} L_{EH}.
\label{eq:ls1}
\end{equation}
Each term is evaluated on samples with the corresponding supervision available.
We train for $450$k updates and set $\lambda_{HW}{=}\lambda_{EH}{=}1.0$ based on validation.
After cross-domain alignment is introduced, UMI, ego-video, and paired ego--UMI batches are sampled in a $6{:}3{:}1$ ratio.
Arm trajectories whose mean relative translational displacement falls below a fixed activity threshold are excluded from Stage~1 training.

Stage~2 transfers the learned representation to the policy through LAM supervision on UMI and robot demonstrations.
The LAM remains fixed, while all VLA parameters $\theta$ and auxiliary query-head parameters $\psi$ are optimized jointly:
\begin{equation}
L_{\mathrm{S2}}
=
L_R+\lambda_A L_{\mathrm{align}}+\lambda_D L_{\mathrm{policy\text{-}dyn}},
\label{eq:ls2}
\end{equation}
where $L_R$ supervises action generation and the auxiliary terms constrain the policy representation using the frozen LAM.
Training runs for $60$k updates.
The first $3$k updates optimize $L_R$ alone; $\lambda_A$ and $\lambda_D$ are then increased linearly from zero to $1.0$ over the next $2$k updates and held constant thereafter.
This schedule introduces representation supervision after the policy has begun adapting to the target demonstrations.

At deployment, the standard $\pi_{0.5}$ flow-matching action expert generates action chunks conditioned on $c_t$, comprising the current visual observations, task instruction, and proprioceptive state.
The LAM, its DINOv2 encoder, and the auxiliary query head are used exclusively during training and are omitted from the deployed policy.
Because the action expert is not conditioned on the auxiliary latent prediction, LAM supervision affects execution through the optimized policy parameters without changing the inference architecture or requiring future observations.

\begingroup
\setlength{\textfloatsep}{0.85\baselineskip plus 0.1\baselineskip minus 0.1\baselineskip}
\setlength{\dbltextfloatsep}{0.85\baselineskip plus 0.1\baselineskip minus 0.1\baselineskip}
\begin{table*}[!t]
  \centering
  \vspace*{4pt}
\begin{minipage}{\textwidth}
  \centering
  \captionof{table}{\textbf{Stage 1 ablations of action grounding and alignment.}
  The full model is compared with Dynamics-Only LAM and variants omitting $L_U$, $L_{HW}$, or $L_{EH}$.
  FDM measures feature predictability; action-decoding errors (right/left, normalized) measure action grounding; HW~R@5 and EH~R@5 measure head--wrist and ego--UMI-head retrieval, respectively.
  UMI validation metrics exclude idle arms; EH retrieval uses a held-out pool ($n{=}1{,}012$, Wilson 95\% CI).
  $^\dagger$: linear action probe evaluated with a frozen backbone.}
  \label{tab:stage1}
  \setlength{\tabcolsep}{4pt}
  \renewcommand{\arraystretch}{1.15}
  \footnotesize
  \begin{tabular}{@{}lcccccc@{}}
    \toprule
    Variant
    & Wrist RMSE $\downarrow$
    & Gripper $\downarrow$
    & Head RMSE $\downarrow$
    & FDM $\uparrow$
    & HW R@5 $\uparrow$
    & EH R@5 $\uparrow$ {[}95\% CI{]} \\
    \midrule
    \textbf{Ours}
      & 0.092 / 0.117
      & 0.082 / 0.068
      & 0.126 / 0.126
      & 0.476
      & 0.180
      & 0.448 {[}0.417, 0.478{]} \\
    Dynamics-Only LAM$^\dagger$
      & 0.295 / 0.373
      & 0.282 / 0.270
      & 0.330 / 0.335
      & 0.589
      & 0.007
      & 0.309 {[}0.282, 0.338{]} \\
    w/o $L_U$$^\dagger$
      & 0.224 / 0.280
      & 0.274 / 0.258
      & 0.254 / 0.260
      & 0.490
      & 0.039
      & 0.390 {[}0.361, 0.421{]} \\
    w/o $L_{HW}$
      & 0.091 / 0.117
      & 0.082 / 0.068
      & 0.128 / 0.130
      & 0.515
      & 0.046
      & 0.137 {[}0.118, 0.160{]} \\
    w/o $L_{EH}$
      & 0.089 / 0.115
      & 0.080 / 0.066
      & 0.121 / 0.122
      & 0.488
      & 0.200
      & 0.328 {[}0.300, 0.358{]} \\
    \bottomrule
  \end{tabular}
\end{minipage}

\end{table*}

\begin{figure}[t]
  \centering
  \includegraphics[width=0.85\columnwidth]{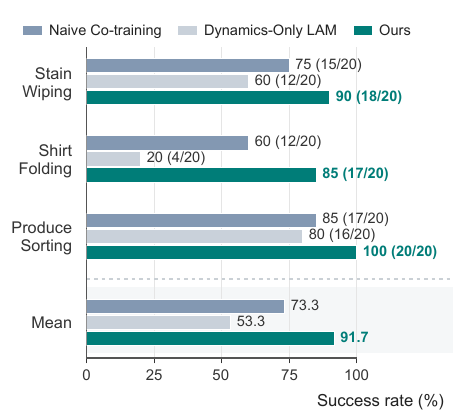}
  \caption{\textbf{Real-robot performance.}
  Success rates on Stain Wiping, Shirt Folding, and Produce Sorting using the full robot demonstration set and the same UMI data.
  Counts in parentheses denote successes over $20$ rollouts per task and policy; Mean averages the three tasks equally.}
  \label{fig:real-robot-performance}
\end{figure}

\begin{figure*}[!t]
  \centering
  \includegraphics[height=1.78in]{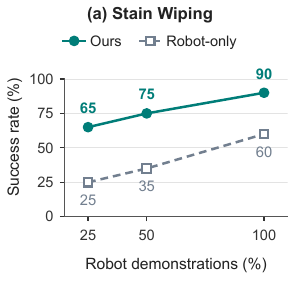}%
  \includegraphics[height=1.78in]{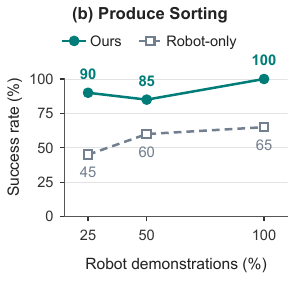}%
  \includegraphics[height=1.78in]{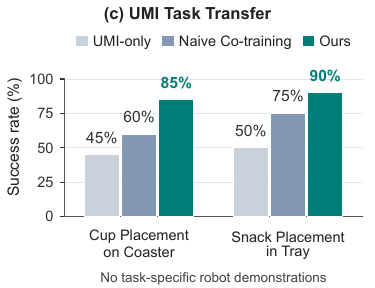}
  \caption{\textbf{Real-robot data efficiency and UMI task transfer.}
  (a,b) Success rates on Stain Wiping and Produce Sorting at three robot demonstration budgets; the UMI dataset and frozen teacher are fixed for Ours.
  (c) Transfer to Cup Placement on Coaster and Snack Placement in Tray using only UMI demonstrations of these tasks.
  Naive Co-training and Ours additionally use robot demonstrations exclusively from other tasks; UMI-only uses UMI data alone.
  Each condition is evaluated over $20$ rollouts.}
  \label{fig:real-robot-results}
\end{figure*}

\section{Experiments}
\label{sec:experiments}

In this section, we demonstrate the effectiveness of UMI-Bridge for action-grounded latent alignment and data-efficient VLA post-training.
We first assess action grounding and alignment across views and domains through Stage~1 ablations.
We then conduct real-robot experiments to address three questions:
\textbf{Q1:} Does LAM regularization improve task success over action-only co-training, and how does teacher quality affect performance?
\textbf{Q2:} How effectively does UMI-Bridge reduce the need for robot demonstrations?
\textbf{Q3:} Can UMI-Bridge transfer behaviors demonstrated only with UMI to real-robot execution without task-specific robot demonstrations?

\subsection{Experimental Setup}
\label{sec:exp-setup}

The real-robot platform is a Tianji Marvin semi-humanoid bimanual robot with two $7$-DoF arms, each fitted with the same 3D-printed UMI gripper design used during UMI data collection, and observed by three RGB cameras---one head-mounted and one on each wrist---at $224{\times}224$.
Our task suite consists of five manipulation tasks divided into two groups.
Three \emph{classic} tasks---\emph{Stain Wiping}, \emph{Shirt Folding}, and \emph{Produce Sorting}---have both robot and UMI demonstrations.
These tasks involve wiping stains from a tabletop, folding a T-shirt, and sorting fruits and vegetables, respectively.

Two \emph{UMI-only} tasks---\emph{Cup Placement on Coaster} and \emph{Snack Placement in Tray}---have UMI demonstrations but no task-specific robot demonstrations.
The transfer tasks contribute only UMI demonstrations; any robot demonstrations used for co-training come from the three classic tasks.
This setting evaluates whether UMI-Bridge can execute UMI-demonstrated behaviors on the robot without robot demonstrations of the target task.

Each reported task--policy pair is evaluated over $20$ rollouts using task-specific binary success criteria shared by all methods.
The robot demonstration budget, use of UMI data, and teacher variant are specified for each experiment below.

\subsection{Stage 1: Action Grounding and Alignment}
\label{sec:exp-stage1}

We evaluate action grounding and correspondence across views and domains in the Stage~1 latent (Sec.~\ref{sec:action-equivalence}).
Table~\ref{tab:stage1} compares the full objective with Dynamics-Only LAM and variants omitting $L_U$, $L_{HW}$, or $L_{EH}$, using the same backbone, data, and training schedule.
We report FDM scores for feature predictability, normalized action-decoding errors for action grounding, and head--wrist (HW~R@5) and ego--UMI-head (EH~R@5) retrieval for correspondence.

Dynamics-Only LAM achieves the highest FDM score ($0.589$ versus $0.476$), but much higher wrist-action RMSE than the full model ($0.295/0.373$ versus $0.092/0.117$).
Removing $L_U$ while retaining both alignment losses also increases wrist RMSE to $0.224/0.280$ and reduces HW~R@5 from $0.180$ to $0.039$.
Visual predictability alone therefore does not ensure action grounding; action supervision improves both action decoding and head--wrist correspondence.

Removing $L_{HW}$ leaves wrist RMSE nearly unchanged ($0.091/0.117$), but reduces HW~R@5 from $0.180$ to $0.046$ and EH~R@5 from $0.448$ to $0.137$, despite retaining $L_{EH}$.
Accurate action decoding within each view thus does not ensure cross-view consistency, and head--wrist alignment also benefits correspondence between egocentric video and UMI head observations.

Without $L_{EH}$, EH~R@5 drops from $0.448$ to $0.328$, while the metrics evaluated within the UMI domain improve slightly.
Explicit alignment on paired human--UMI clips therefore improves cross-domain correspondence beyond a shared backbone and UMI action supervision, with a modest trade-off in UMI validation performance.

These ablations support combining action grounding with cross-view and cross-domain alignment to learn action-relevant latents.
The full model serves as the frozen Stage~2 wrist teacher; Sec.~\ref{sec:exp-main} compares it with the Dynamics-Only LAM teacher to assess the effect on real-robot policy performance.

\subsection{Real-Robot Experiments I: Policy Performance}
\label{sec:exp-main}

We compare three VLA post-training variants using the full robot demonstration set and the same UMI data.
\emph{Naive Co-training} optimizes action supervision alone, without latent regularization.
\emph{Dynamics-Only LAM} applies the Stage~2 regularization losses using a LAM teacher trained only with $L_{\mathrm{dyn}}$.
\emph{Ours} uses the full action-grounded LAM as its teacher.
Both LAM-regularized variants use the same regularization losses and weights.
Fig.~\ref{fig:real-robot-performance} reports results on Stain Wiping, Shirt Folding, and Produce Sorting.

Ours achieves the highest observed success rate on all three tasks, with a mean of $91.7\%$ versus $73.3\%$ for Naive Co-training and $53.3\%$ for Dynamics-Only LAM---gains of $18.3$ and $38.3$ percentage points, respectively.
The improvement over Naive Co-training is present in Stain Wiping ($75\%\rightarrow90\%$), Shirt Folding ($60\%\rightarrow85\%$), and Produce Sorting ($85\%\rightarrow100\%$), indicating that the benefit extends across the three evaluated tasks.

Dynamics-Only LAM performs worse than Naive Co-training on every task, with the largest gap on Shirt Folding: $20\%$ versus $60\%$, while Ours reaches $85\%$.
This reverses the ranking by forward-dynamics score in Stage~1, where Dynamics-Only LAM scores higher than the full teacher ($0.589$ versus $0.476$) despite substantially worse wrist-action decoding.
Thus, better visual predictability alone does not identify a useful regularization target for control.
The real-robot comparison supports the full action-anchored teacher design; the individual losses are examined separately in the Stage~1 ablations.

\begin{figure}[t]
  \centering
  \includegraphics[width=0.8\columnwidth]{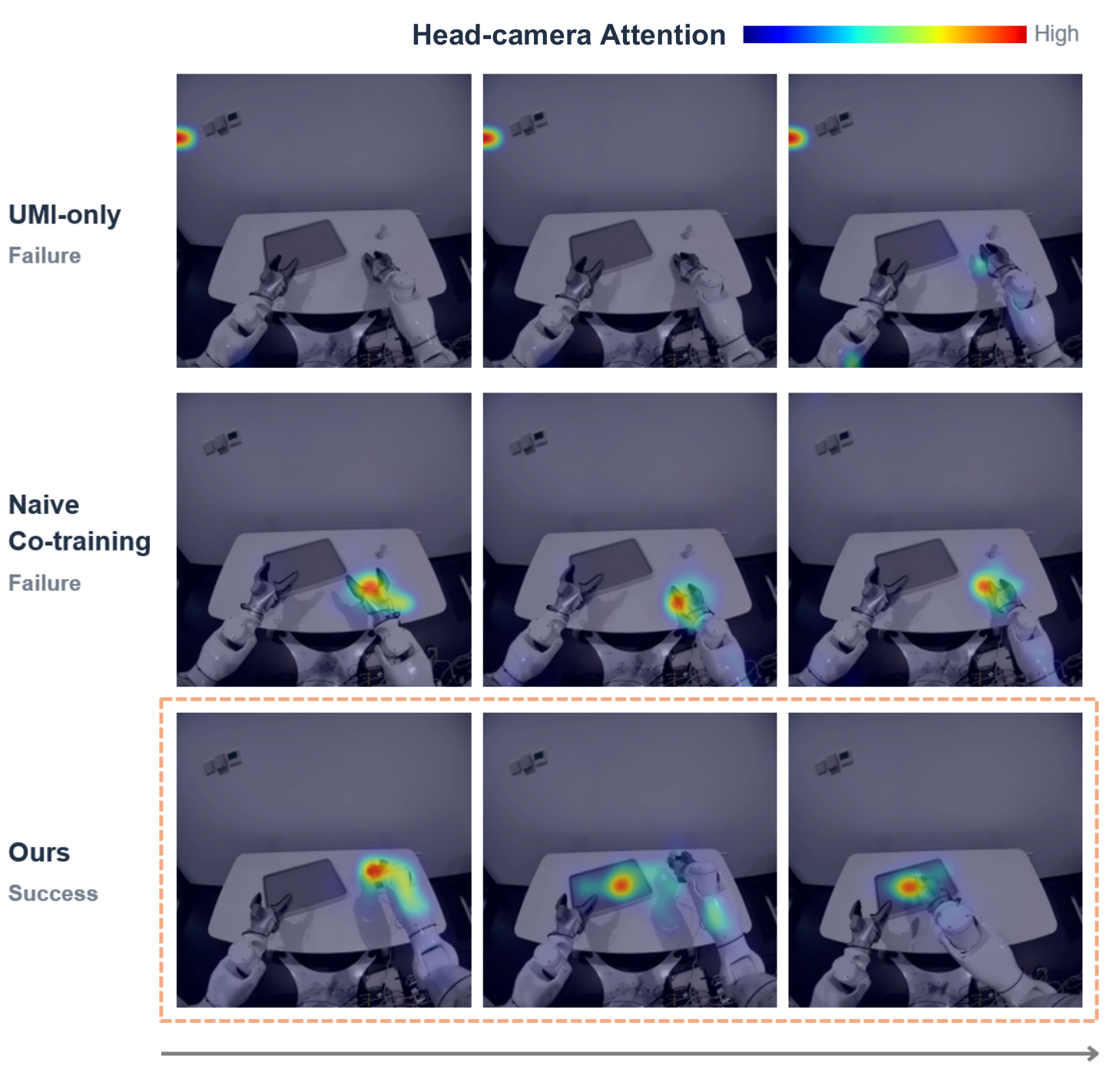}
  \caption{\textbf{Head-camera attention during real-robot execution.}
  Selected frames from Snack Placement in Tray are shown for UMI-only, Naive Co-training, and Ours (top to bottom).
  We visualize action-query attention to head-camera image tokens at L9/H3 in each model, averaged over $10$ denoising steps and $16$ action queries.
  Maps are independently min--max normalized, upsampled with bicubic interpolation, and lightly smoothed for display.
  Warmer colors indicate higher relative attention within each image; arrows indicate chronological order.}
  \label{fig:head-attention}
\end{figure}

\subsection{Real-Robot Experiments II: Data Efficiency}
\label{sec:exp-data-eff}

We evaluate how performance changes with the amount of robot demonstration data on \emph{Stain Wiping} and \emph{Produce Sorting}.
Ours and Robot-only are trained at robot demonstration budgets of $25\%$, $50\%$, and $100\%$, using the same training schedule.
For Ours, the UMI dataset and frozen teacher are held fixed across budgets; Robot-only uses robot action supervision without the UMI channel or latent regularization.
Each task--method--budget combination is evaluated over $20$ rollouts, with per-task results shown in Fig.~\ref{fig:real-robot-results}(a,b).

At matched robot-data budgets, Ours achieves higher success than Robot-only on both tasks at all three budgets.
Averaged equally over the two tasks, Ours attains $77.5\%$, $80.0\%$, and $95.0\%$ success at $25\%$, $50\%$, and $100\%$ robot data, respectively, compared with $35.0\%$, $47.5\%$, and $62.5\%$ for Robot-only.
The corresponding gains are $42.5$, $32.5$, and $32.5$ percentage points. %

With only $25\%$ of the robot demonstrations, Ours reaches $77.5\%$ mean success, exceeding the $62.5\%$ of Robot-only trained on the full dataset by $15.0$ percentage points.
The comparison holds on both tasks: $65\%$ versus $60\%$ on Stain Wiping and $90\%$ versus $65\%$ on Produce Sorting.
This advantage is also reflected in selected Produce Sorting executions with angled object placements: Ours retains the vegetable through lifting and completes both placements, whereas full-data Robot-only repeatedly attempts the grasp but leaves both objects on the table.
Thus, on these two tasks, our full method achieves higher observed success with $75\%$ fewer robot demonstrations while retaining the UMI data.

\subsection{Real-Robot Experiments III: UMI Task Transfer}
\label{sec:exp-transfer}

We evaluate the two transfer tasks in Fig.~\ref{fig:real-robot-results}(c) without task-specific robot demonstrations.
\emph{UMI-only} is post-trained on UMI data alone; Naive Co-training and Ours also use robot demonstrations from other tasks.
Naive Co-training and Ours use matched training data, with Ours additionally applying the frozen LAM regularizer.
All three policies use head and wrist observations.

On \emph{Cup Placement on Coaster}, Ours achieves $85\%$ success, compared with $60\%$ for Naive Co-training and $45\%$ for UMI-only.
On \emph{Snack Placement in Tray}, the corresponding rates are $90\%$, $75\%$, and $50\%$.
Averaged equally over both tasks, Ours achieves $87.5\%$ success, exceeding Naive Co-training and UMI-only by $20$ and $40$ percentage points, respectively.
These results support LAM regularization for UMI-to-robot task transfer.

Fig.~\ref{fig:head-attention} visualizes attention maps extracted from head-camera images during policy execution on Snack Placement in Tray.
In the selected rollout, Ours concentrates attention on the manipulated object as it is grasped and transferred to the tray, whereas the baselines exhibit less consistent focus on the target object.
This object-centered attention is consistent with the intended role of action-anchored latent supervision: guiding the policy toward visual cues that matter for manipulation.
Together with the higher task success rate, these visualizations provide qualitative support for transferring action-relevant representations through the UMI bridge.

\endgroup

\section{Conclusion}
\label{sec:conclusion}

We presented UMI-Bridge, an action-anchored latent alignment framework for VLA training on UMI and robot demonstrations.
Its frozen, action-grounded LAM regularizes VLA post-training without changing the inference architecture.
Experiments show that visual predictability alone does not ensure effective policy supervision, while our method improves task success, robot-data efficiency, and transfer from UMI demonstrations without task-specific robot data.
In future work, we are interested in investigating how the Ego-to-UMI data ratio and the diversity of tasks, objects, and scenes in each source affect policy success and generalization.
We believe these studies will characterize when egocentric experience complements UMI supervision and guide data selection for robot learning.

\bibliographystyle{IEEEtran}
\bibliography{refs}

\end{document}